\documentclass[11pt,a4paper]{article}
\usepackage[hyperref]{acl2021}
\usepackage{times}
\usepackage{latexsym}

\usepackage{graphicx}
\usepackage{microtype}
\usepackage{comment} 

\usepackage{soul}

\aclfinalcopy 

\title{KnowGraph@IITK at SemEval-2021 Task 11: Building Knowledge Graph for NLP Research}

\author{
    Shashank Shailabh$^{*}$ \qquad   
    Sajal Chaurasia\thanks{\quad Authors equally contributed  to this work.} \qquad
  \large{\textbf{Ashutosh Modi}} \\
{Indian Institute of Technology Kanpur (IIT Kanpur)} \\
{\tt \{shailabh, sajal\}@iitk.ac.in}\\
  {\tt ashutoshm@cse.iitk.ac.in}  \\
}

\date{}

\begin{document}
\maketitle
\begin{abstract}
Research in Natural Language Processing is making rapid advances, resulting in the publication of a large number of research papers. Finding relevant research papers and their contribution to the domain is a challenging problem. In this paper, we address this challenge via the  SemEval 2021 Task 11: NLPContributionGraph, by developing a system for a research paper contributions-focused knowledge graph over Natural Language Processing literature. The task is divided into three sub-tasks: extracting contribution sentences that show important contributions in the research article, extracting phrases from the contribution sentences, and predicting the information units in the research article together with triplet formation from the phrases. The proposed system is agnostic to the subject domain and can be applied for building a knowledge graph for any area. We found that transformer-based language models can significantly improve existing techniques and utilized the SciBERT-based model. Our first sub-task uses Bidirectional LSTM (BiLSTM) stacked on top of SciBERT model layers, while the second sub-task uses Conditional Random Field (CRF) on top of SciBERT with BiLSTM. The third sub-task uses a combined SciBERT based neural approach with heuristics for information unit prediction and triplet formation from the phrases. Our system achieved F1 score of 0.38, 0.63 and 0.76 in end-to-end pipeline testing, phrase extraction testing and triplet extraction testing respectively.
\end{abstract}
\section{Introduction}
Given the advancements in Natural Language Processing (NLP), a large number of research papers are published every year. However, given the field's dynamic nature, keeping track of all the papers is a non-trivial task. This motivated the formulation of an Open Research Knowledge Graph \cite{jaradeh2019open}, a knowledge graph of research contributions and the relation between them. Task 11 of SemEval 2021  \cite{ncg} formalizes the building of a contributions-focused knowledge graph of NLP literature.
The task is divided into three sub-tasks:
\begin{itemize}
    \item \textbf{Sub-task A} Extracting sentences that posit contributions in a research paper.
    \item \textbf{Sub-task B} Extracting relevant phrases that include scientific terms and relational cues from the extracted sentences of the sub-task A.
    \item \textbf{Sub-task C} Triplet (subject phrase, predicate phrase, object phrase) formation from the extracted phrases of the sub-task B and classification of the triplet in one of the information units (IU). There are twelve information units (Research problem, Approach, Results, Model, Code, Dataset, Experimental setup, Hyperparameters, Baselines, Tasks, Experiments, and Ablation analysis), each focusing on different sections in a research paper. These information units can represent all the findings given in a research paper. Out of twelve, first three information units are present in each article. 
\end{itemize}
\begin{figure*}[t]
\centering
\includegraphics[width=\textwidth,height=2.8in]{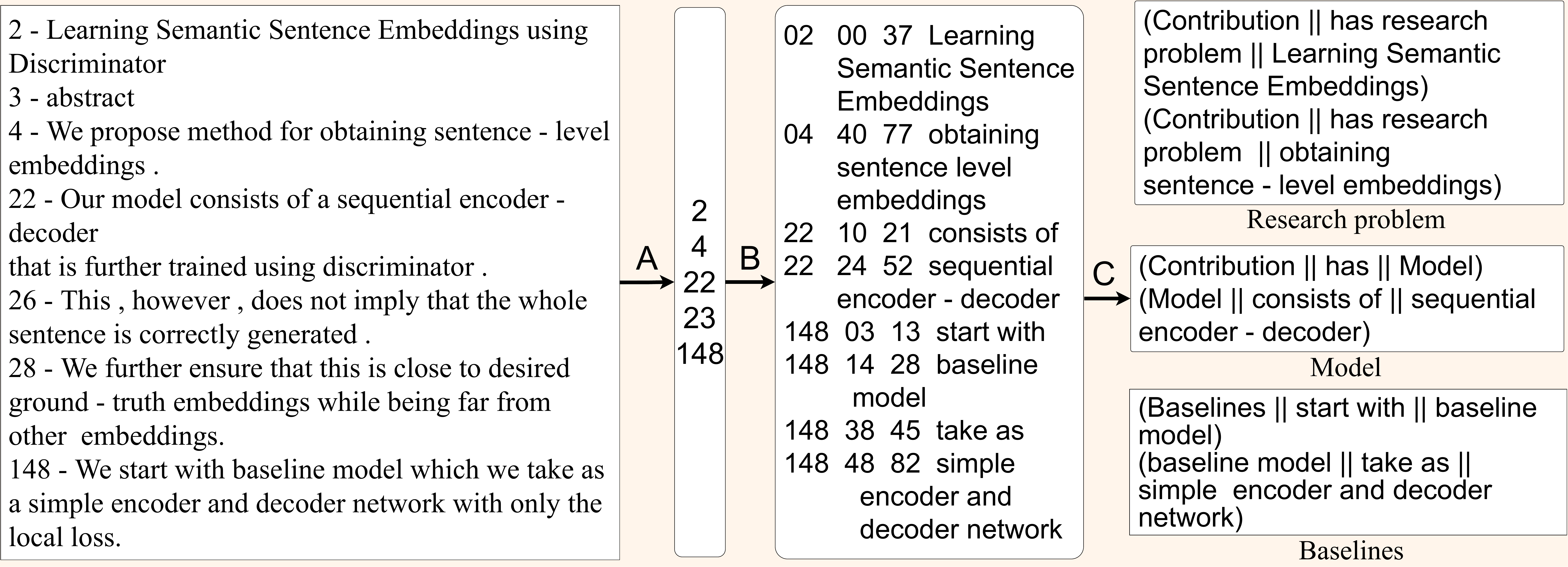}
\caption{Dataset Example} 
\label{fig:System-exam}
\end{figure*}
Recently, transformer-based approaches have been popular for NLP applications \cite{liu-lapata-2019-text, 2019arXiv190903193Y}. For the sub-task A, we propose a sentence level classifier, leveraging SciBERT \cite{beltagy-etal-2019-scibert} and Bidirectional Long Short-Term Memory (BiLSTM) \cite{hochreiter1997long}. SciBERT model has been trained on 1.14M scientific papers from Semantic Scholar corpus, which has 18\% papers from the computer science domain. Our system for sub-task B also uses SciBERT based model using CRF \cite{lafferty2001conditional} on top of BiLSTM layers using BILUO (B=start token of phrase, I=interior tokens of phrase, L=last token of phrase, U=single token phrase and O=Non-phrase tokens) labelling scheme for tokens. For sub-task C, we build a combination of neural and heuristic-based approach. The IU prediction for sub-task C is at document level where two information units use heuristic approach while others use a multi-label classifier based on BiLSTM stacked on top of SciBERT. For triplet formation in sub-task C, we use a separate SciBERT+BiLSTM classifier along with heuristics. These triplets are classified into one of the already predicted information units, i.e., the output unit having the maximum score among the predicted IU. However, some of the IU, such as Baselines, Ablation analysis, Code and Research problem triplets, perform well using heuristic, so we use heuristic instead of a neural approach for these IU.
Our proposed system was ranked third in the overall end-to-end pipeline (A+B+C) testing and achieved F1 score of 0.38. Our proposed model was ranked fourth in phrase extraction (A\textsubscript{GT}+B+C) and triplet extraction testing (A\textsubscript{GT}+B\textsubscript{GT}+C) with a F1 score of 0.63 and 0.76 respectively where A\textsubscript{GT} and B\textsubscript{GT} represent ground-truth for sub-task A and B respectively. Phrase extraction testing uses ground-truth labels for sub-task A, while triplet extraction testing uses ground-truth for both sub-task A and B. We found that the heuristic-based model for two IU (Research problem and Code) in sub-task C can achieve high performance and achieved F1 score of 0.98 and 1.00, respectively. Our system implementation code is made available via GitHub\footnote{\url{https://github.com/sshailabh/SemEval-2021-Task-11}}.
\section{Background}
\subsection{Problem Definition}
Consider a document $D =\{s_{1},s_{2},..,s_{i},..,s_{N}\}$ having $N$ sentences $s_{i}$. Sub-task A finds $M$ contribution sentences denoted by $S = \{s_{1},s_{2},...,s_{M}\}$ from $D$. Sub-task B selects phrases $P=\{p_{1},p_{2},..,p_{i},..,p_{L}\}$ where $p_{i}$ is a phrase selected from a sentence $s \in S$ and $L$ is total number of phrases in $D$. Sub-task C is forming triplets of extracted phrases for IU denoted by $U =\{u_{1},..,u_{i},..,u_{X}\}$ where $u_{i}$ is one of the twelve IU and $X$ is number of IU in document $D$ ranging between three to twelve. For each $u_{i}\in U$, there is a triplet set called $T^{i} = \{({su}_{1}^{i},{pr}_{1}^{i},{ob}_{1}^{i}),({su}_{2}^{i},{pr}_{2}^{i},{ob}_{2}^{i}), ..,({su}_{j}^{i},{pr}_{j}^{i},{ob}_{j}^{i})\\
,..,({su}_{O}^{i},{pr}_{O}^{i},{ob}_{O}^{i})\}$ where $({su}_{j}^{i},{pr}_{j}^{i},{ob}_{j}^{i})$ is a triplet representing subject, predicate(relation) and object respectively and $O$ is total number of triplets in $u_{i}$ IU in document $D$. An example dataset is given in Figure \ref{fig:System-exam} for reference. Here, on moving from left to right is research paper, contribution sentences, phrases and IU along with triplets given in a research paper respectively.

\subsection{Related Work}
Knowledge graphs \cite{rebele2016yago, hertling2018dbkwik, lehmann2015dbpedia, carlson2010coupled} have shown to be helpful in several areas such as search, knowledge extraction, inter alia. However, only a handful are based on research articles \cite{pmid32591513}. Typically, most knowledge graphs are created with rule-based approaches, hence, limiting their performance and generalization. However, some recent approach such as \citet{sang2018sematyp, wang2020covid} uses neural approach in biomedical literature. To the best of our knowledge, there is no available contributions-focused knowledge graph over NLP literature using the neural approach.

\noindent \textbf{Sub-task A:} The sub-task of extracting contribution sentences can also be posed as an extractive summarization problem  \cite{nallapati2016summarunner, narayan-etal-2018-ranking, liu2019fine, cheng-lapata-2016-neural, zhou-etal-2018-neural-document, dong-etal-2018-banditsum, wang-etal-2020-heterogeneous}. BERTSUM \cite{liu-lapata-2019-text} and MATCHSUM \cite{zhong-etal-2020-extractive} are the recent methods leveraging language models and uses ROUGE-1, ROUGE-2 and ROUGE-L scores \cite{lin-2004-rouge} on DailyMail data-set \cite{NIPS2015_5945}. However, this extractive summarization technique may not be applicable in our case due to a number of reasons. Firstly, extractive summarization alone will not give all the contribution sentences because some sentences may not be relevant to the summarization task. Secondly, extractive summarization models are not tested on large documents such as research articles due to the limitation of the input token length for transformer-based language models. Some long document transformer-based methods are proposed (e.g., \citealp{beltagy2020longformer}), and can consider documents up to a length of $4096$ tokens, however, in our case, documents have on an average $\sim$10,000 tokens. Some of the extractive summarization methods \cite{liu-lapata-2019-text, miller2019leveraging} take the number of contribution sentences as a hyper-parameter, but in our case, this is a trainable parameter in our model.

\noindent \textbf{Sub-task B:} Sub-task B closely resembles the phrase extraction problem and several neural methods (\citet{pmid32413094, wang2016ptr, zhang-etal-2016-keyphrase}, inter alia) and non-neural based methods (using n-grams and noun-phrases with certain Part-of-speech (POS) patterns \cite{hulth-2003-improved}) have been proposed. \citet{gollapalli2017incorporating} have shown that CRF has the potential to improve the existing phrase extraction model. \citet{alzaidy2019bi} jointly leverages CRF and BiLSTM to capture hidden semantics for phrase extraction. \citet{pmid32413094} extended the work of \citet{alzaidy2019bi} with the idea of self-training and used word embeddings, POS embeddings, and dependency embeddings with a BILUO labelling scheme in the output.  Our proposed model took inspiration from \citet{pmid32413094} and propose a SciBERT based model using CRF on top of BiLSTM layers using BILUO labelling scheme on tokens. Our model captures better semantics than the word embeddings based approach in \citet{pmid32413094} because of SciBERT, which is trained on the scientific corpus. Moreover, our model uses the WordPiece tokenizer and hence, robust to Out-of-Vocabulary (OOV) tokens. \citet{sahrawat2020keyphrase} used contextual embeddings to the BiLSTM and CRF model using BIO (B=start token of phrase, I=continuation tokens of phrase and O=Non-phrase tokens) labelling scheme. \citet{ratinov-roth-2009-design} discussed that the BILUO scheme is superior to the BIO scheme; hence we adopt the BILUO scheme for sub-task B. Recently, \citet{lai-etal-2020-joint} combined sequence labelling with joint learning inspired from self-distillation to boost model performance on unsupervised datasets. However, their model used BIO labelling scheme and gave a comparable or marginal improvement in a supervised setting.

\noindent \textbf{Sub-task C:} The sub-task C can be divided into two parts - information units (IU) prediction and triplet formation. The information unit prediction and triplet formation have been approached in the literature mainly using rule-based methods. \citet{rusu2007triplet} suggests using syntactic parsers for generating parse trees, followed by triplet extraction using parser dependent techniques. \citet{jivani2011multi} proposed an algorithm that exhibits the relationship between subject and object in a sentence using Stanford parser. This rule-based algorithm can form multiple triplets from a sentence as compared to \citet{rusu2007triplet}. Stanford OpenIE Relation triplet formation \cite{angeli-etal-2015-leveraging} uses a classifier, which learns to extract self-contained clauses from longer sentences to form the final triplets using heuristics. \citet{Hamoudi2016ExtractingRT} and \citet{Jaiswal2015AMA} are also rule-based methods for triplet formation using Stanford dependency parser and constituency parser, respectively. KG-Bert \cite{2019arXiv190903193Y} uses the BERT language model and utilize entity and relation descriptions of a triplet to compute its scoring function.
\section{System Overview}
\begin{figure*}[t]
\centering
\includegraphics[width=0.95\textwidth,height=4.3in]{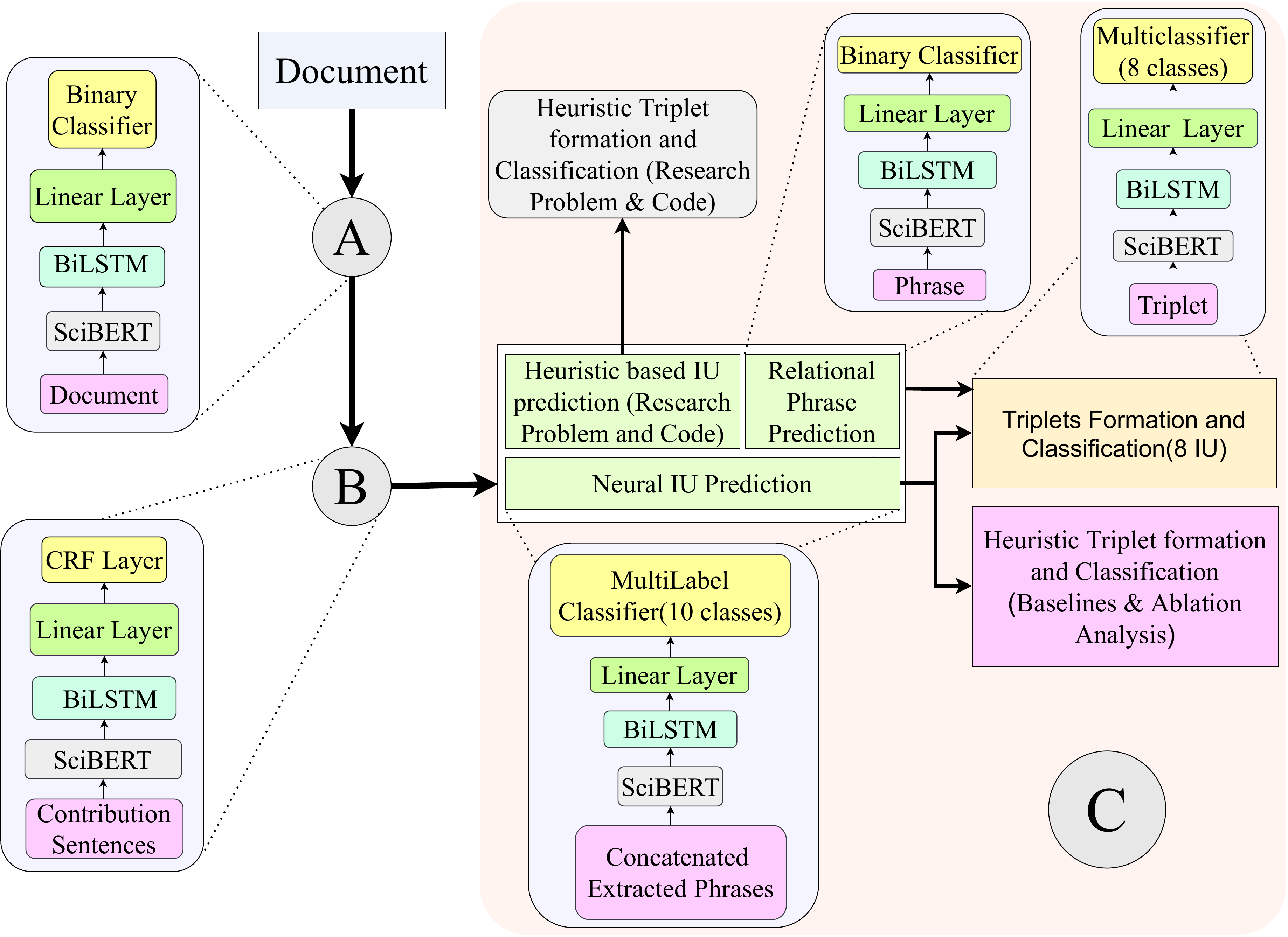}
\caption{ Overall system architecture for end-to-end pipeline showing all the sub-tasks and their respective model.}
\label{fig:system-arch}
\end{figure*}
The proposed system is shown in Figure \ref{fig:system-arch} depicting the entire pipeline and its respective model.
\subsection{Sub-Task A}
\noindent \textbf{Initial Experimentation:} We experimented with SciBERT, a language model trained on research papers to build a binary classifier. The classifier encodes the sentence using feature representation corresponding to [`CLS'] token from the last layer of the pre-trained SciBERT model to predict a binary label. Since the fine-tuning dataset is small, linear layers were not able to capture contextual information well. We experimented by adding a Convolution Neural Network \cite{zhang2015sensitivity} layer on the top of SciBERT. The CNN architecture use three kernels of size two, three and four. The model boosted the performance; however, the model cannot classify long sentences due to their ambiguous nature and lack of CNN's capacity to capture long semantic dependency among tokens.

\noindent \textbf{Proposed Approach:} We propose the SciBERT+BiLSTM model (a sentence-level binary classifier where BiLSTM is stacked on top of the SciBERT model). This helps to encode hidden semantics and long-distance dependency. Consider the training dataset as $Tr =\{D_{1},D_{2},..,D_{i},..,D_{Z}\}$ comprising of $Z$ documents. Each $D_{i}$ can be represented as $D_{i} =\{s_{i1},s_{i2},..,s_{ij},..,s_{iN}\}$ where $N$ is the number of sentences in the document and $s_{ij}$ is the $j^{th}$ sentence of document $D_{i}$. Each sentence is assigned a ground-truth label where label ``1" represents a contribution sentence and label ``0" a non-contribution sentence. The sentences are processed using a SciBERT model followed by stacked BiLSTM layers whose output is further processed through linear layers with ReLU \cite{icml2010_103} non-linearity. We add dropout layers to avoid overfitting. The last linear layer consists of two units corresponding to label ``0" and label ``1". The final output label is the label whose corresponding unit has a higher score in the last linear layer. Our loss function is weighted binary cross-entropy loss, where weights are in according to the number of samples in each class.
\subsection{Sub-Task B}
\noindent \textbf{Initial Experimentation:} We built our initial method, the BiLSTM+CRF model, on the lines of \citet{pmid32413094}. The model uses a BiLSTM layer along with CRF for sequence to sequence (BILUO) labelling in order to mark the phrases in a sentence.
We introduced SciBERT (by replacing BiLSTM layers) to fine-tune and better generalise with an increase in performance. We tested the significance of CRF by replacing it with a softmax layer, which gave poor performance since it is unable to learn the constraints in the BILUO scheme.

\noindent \textbf{Proposed Approach:} We further improved the SciBERT+CRF model to improve semantic information. Our proposed model stack BiLSTM layers on top of SciBERT, followed by CRF (see Figure \ref{fig:system-arch}).
 The word-level representation $\{x_1,x_2,...,x_N\}$ from the input sentence passes through the SciBERT tokenizer. We used the representation of the first sub-token for every word as the input to the SciBERT. The tokenized input is passed into the SciBERT layer, followed by the BiLSTM layer. The final feature output is mapped to a hidden linear layer to get the score matrix $Z$, which is passed into the CRF layer for label prediction $y$. The CRF layer is the same as described in \citet{lample2016neural}.

The output produced by the SciBERT + BiLSTM + Hidden Layer corresponds to a scoring matrix ${Z^{(n \times l)}}$ where $n$ denotes the number of words in the input sentence, and $l$ is the number of labels ($l$=5). The score of an output sequence ${y}$ using CRF is given by:
\begin{equation}
    Scr(s,y) = \sum_{i=0}^{n}(Z_{i,y_i} + T_{y_{i-1},y_i})
\end{equation}
where $Z_{i,j}$ denotes the score of word ${w_i}$ with the $j_{th}$ label,  $T_{y_{i-1},y_i}$ is the transition score from the label $y_{i-1}$ to $y_i$,
$y = \{y_1,y_2,....,y_n\}$ is the sequence of true labels and $Scr(s,y)$ corresponds to output score for sentence $s$ and true labels $y$. A softmax over all possible label sequences yields a probability for true labels $y$:
\begin{equation}
P(y|s) = \frac{exp(Scr(s,y))}{\sum_{y' \in Y(s)}^{}exp(Scr(s,y'))}    
\end{equation}
where Y(s) corresponds to all the possible label sequences for sentence $s$. 
Now during training, our task is to maximize the log-probability of the correct label sequence $y$. 
Model loss is defined as follows:
\begin{equation}
    {\small} L(\Theta) = -(1/M)\sum_{i=1}^{M}log(P(y_i|s_i)) + \frac{\lambda}{2}\left \| \Theta  \right \|^2
\end{equation}    
where $s_i$ is the input sentence, $y_i$ is the corresponding true label sequence, $\Theta$ denotes the model parameters, $\lambda$ is the regularization hyperparameter and M is the train set size. Output label prediction is made by:
\begin{equation}
    y^* =  {argmax_{y'\in Y(s)}}P(y'|s) 
\end{equation}
Here ${y^*}$ represents the final output label sequence, $s$ is the input sentence, $Y(s)$ is the set of possible label sequences and $P(y'|s)$ denotes the probability of getting $y'$ label sequence from sentence $s$. The phrases are extracted using BILUO scheme based on the prediction outputs.

\subsection{Sub-Task C}
\noindent \textbf{Initial Experimentation:} We used a combination of neural and rule-based approach for sub-task C. An IU triplet has three phrases - subject, predicate and object. Research problem and Code IU triplets have fixed subject and predicate in triplets. We employed a heuristic to scan the phrases of the first thirty lines of each document and select only those sentence's phrases that have only a single phrase extracted out. These phrases form object in Research problem IU triplets. A regex expression is used to extract all the sentences in the article that contain any URLs for Code IU triplet's object. 
However, only those URL sentences are selected which have token such as ``our" or ``code" or ``our code".

\noindent Our initial approach was to form the triplets for all other IU and classify them into one of the ten remaining information units using SciBERT + BiLSTM multi-class classifier (BiLSTM layers stacked on top of SciBERT). The heuristic for triplet formation is based on orthographic visualization of the document - firstly, the phrases are arranged in the exact order as they appear in the original sentence. Then, every three consecutive pair of phrases present within the same sentence are considered as one triple. This approach gave us decent results since most of the research paper is written in the active voice; hence the subject phrase should occur first, then its corresponding predicate and last should be the object phrase. The SciBERT + BiLSTM multi-class classifier takes concatenated triplets as the input and their corresponding information unit as the ground truth. The loss is a cross-entropy loss, with the final softmax layer having ten classes corresponding to 10 information units. One of the drawbacks of the model is that the IU prediction depends on triplet formation's correctness. Further, a single triplet does not have enough context to be correctly classified into the correct information unit. We visualized triplet formation for feature extraction and found that some IU triplet's, such as Baselines and Ablation analysis, can be better formed using heuristics. Our proposed approach is a better triplet formation model and eliminates these limitations.

\noindent \textbf{Proposed Approach:} Our proposed approach has some information unit such as Research problem and Code, whose model for prediction and triplet formation is entirely heuristic-based and the same as initial experimentation. Our proposed method for the rest of ten IU is divided into two parts - 

\noindent \textbf{IU Prediction -} We propose a SciBERT+BiLSTM multi-label classifier (BiLSTM layers stacked on top of SciBERT) (refer to Figure \ref{fig:system-arch}) whose input is the concatenated phrases and predicts the IU of the document.
The concatenation is in the order of the occurrence of phrases in the document.
Moreover, these concatenated extracted phrases represent the whole document since it includes all relevant keywords of the research article necessary for information unit prediction. Hence, our proposed model encodes information at the document level, which makes it superior to the initial experimentation method.

\noindent \textbf{Triplet Formation and Classification -} In general, a phrase is either a relational phrase or a scientific term (Figure \ref{fig:System-exam}). A specific trend observed in ground truth triplets is that a triplet's predicate is unique for a triplet. This contrasts with the subject and object phrase, which can be used multiple times in other triplets. Hence, a one-to-one relation between predicate phrase and triplet exists.
We identify all predicate phrases from the extracted phrases of sub-task B; then, a corresponding triplet will be formed for each predicate phrase. We train a SciBERT+BiLSTM based binary classifier (BiLSTM layers stacked on top of SciBERT model) to identify the phrases which act as predicates (relational phrases). The model is fine-tuned on our dataset and uses the weighted binary cross-entropy loss. In labelling, ``1" denote as predicate while ``0" denote as non-predicate. To form triplets, we use a simple heuristic to arrange the phrases in the exact order as they appear in the original sentence. For every phrase predicted as a predicate, we take its previous phrase as the subject and its next phrase as the triplet object.

Now, a multi-class classifier for triplet classification for 8 IU (corresponding to all IU except Research problem, Code, Baselines and Ablation analysis) (refer to Figure \ref{fig:system-arch}) is built, which is similar to the one described in the initial experimentation. Since we have already predicted the information units, during inference, the triplet can be assigned only to one of the already predicted information units, i.e., the output unit having the maximum score among the predicted IU.

We used rule-based heuristics for triplet formation of Baselines and Ablation analysis IU. The target sentences, whose phrases belong to baselines IU, are identified by selecting all the headings (i.e. lines having no punctuation) with words such as ``baseline", ``comp" using a regex expression. Then, we took all the sentences between selected headings and their consecutive headings as the target sentences.
The phrases associated with extracted sentences are used for triplet formation via the rule of three consecutive phrases present within the sentence.
The same method is followed for Ablation analysis IU triplets with only change that the relevant headings are found using the regex expression that identifies if that heading contains the word such as ``ablation", ``analysis".

\section{Experimental Setup}
\subsection{Data}
The dataset annotation scheme is as per \citet{DSouza2020NLPContributionsAA}.
\begin {table}[h]
\centering
\begin{tabular}{p{2.3cm}p{4cm}}
\hline
Token Length & \% of sentences less than token length\\
\hline 
50 & 94.57\\
100 & 99.74\\
150 & 99.93\\
200 & 99.96\\
\hline
\end{tabular}
\caption{Token length statistics on train set}
\label{table:token_stat}
\end {table}
The pre-processed dataset consists of 287 annotated NLP research documents in the English language with ground truth for each sub-task. Train, dev and test set have 237, 50 and 155 documents, respectively.
\begin {table}[h]
\begin{tabular}{p{4cm}|p{1.2cm}|p{1.1cm}}
\hline
Statistics & Train & Dev\\
\hline 
\# Documents & 237 & 50\\
\# Contribution sentences & 5096 & 1032\\
\# Non-contribution sentences & 50105 & 10451\\
\# Avg. Sentences in doc. & 232.915 & 229.7\\
\# Avg. Tokens in sentence & 20.622 & 21.06\\
\# Avg. Contribution Sentences in doc. & 21.38 & 20.24\\
\# Avg. Phrases in doc. & 128.53 & 92.52\\
\# Avg. IU in doc. & 4.43 & 4.46\\
\# Max Tokens in sentence & 396 & 193\\
\hline
\end{tabular}
\caption{Dataset Statistics}
\label{table:data_stat1}
\end {table}
We have chosen 100 as the maximum token length in a sentence with WordPiece tokenizer since 99.7\% sentences in the train set have less than or equal to 100 tokens. The Table \ref{table:token_stat} shows token length and percentage(\%) of sentences less than that length. In the sub-task C, if there is no suitable predicate available in the extracted phrases, then the triplet's predicate is chosen from the predefined set of predicates, i.e. ``has", ``on", ``by", ``for", ``has value",``has description", ``based on", ``called". Table \ref{table:data_stat1} shows the dataset statistics related to sub-task A and B. 
\begin {table}[h]
\begin{tabular}{p{4.1cm} p{1.2cm} p{1.2cm}}
\hline
\# Information Unit Triplets & Train & Dev\\
\hline 
Research problem & 635 & 164\\
Approach & 529 & 233\\
Model & 3548 & 570\\
Code & 40 & 9\\
Dataset & 240 & 8\\
Experimental setup & 1928 & 302\\
Hyperparamters & 2267 & 254\\
Baselines & 1625 & 146\\
Results & 4989 & 657\\
Tasks & 0 & 277\\
Experiments & 1472 & 149\\
Ablation analysis & 1407 & 155\\
\hline
\end{tabular}
\caption{Number of triplets in each Information Unit on train and dev set.}
\label{table:data_stat2}
\end {table}
The dataset statistics for sub-task C is given in Table \ref{table:data_stat2}. Tasks information unit has no triplets in train set while Code and Dataset information units have very few triplets in dev set.

\subsection{Hyperparameters}
We have used the dev set to tune our hyperparameters. In every neural model, we are fine-tuning SciBERT.\footnote{\url{https://github.com/huggingface/transformers}} We tried different batch sizes and learning rates for fine-tuning \cite{dodge2020fine}. We found the best results using the AdamW optimizer in the neural models.\\
\textbf{Sub-task A} : We used batch size = 32, learning rate = 1e-05, epoch = 2, two layers of BiLSTM with hidden dimension of 400 and three linear layers (size = 800, 400, 100) with dropout = 0.1. Oversampling of minority class (contribution sentences) counters the skewness in data.\\
\textbf{Sub-task B} : We used batch size = 1, learning rate = 2e-05, epoch = 4, single layer of BiLSTM with hidden dimension = 200 and linear layer with CRF.\\
\textbf{Sub-task C} : We used batch size = 4, max-tokens = 512, learning rate = 2e-05, epoch = 16, threshold on sigmoid output = 0.5, two layers of BiLSTM with hidden dimension of 400 and three linear layers (size = 800,400,100) with dropout = 0.2 for multi-label classification of information units (ten out of twelve). 
We have used batch size = 32, max tokens = 25, learning rate = 2e-05, epoch = 4, two BiLSTM layers with hidden dimension = 400 and three linear layers (size = 800, 400, 100) with dropout = 0.1 for relational phrase prediction model. We have used batch size = 16, max tokens = 50, learning rate = 2e-05, epoch = 2, two BiLSTM layers with hidden dimension = 400 and three linear layers (size = 800, 400, 100) with dropout = 0.2 for triplet classification (eight out of twelve info-units).

\subsection{Evaluation Metrics}
In this task, organizers used Precision, Recall and F1 score metrics. In sub-task A, the predicted and ground-truth contribution sentences of the document calculate the metrics score. The sub-task B output has predicted phrase, contribution sentence number, starting and ending character number of the predicted phrase (Figure \ref{fig:System-exam}). These outputs are the basis for sub-task B metric calculations.
The sub-task C has two groups of metrics. First is the Information Units prediction of a document (regardless of triplets). The second group calculates using both Information Units and triplets (the predicted triplet is correct only if it exactly matches the ground truth triplet and the ground-truth IU; otherwise, it is incorrect). The final score is the average of all the four F1 scores on the dataset. The participating team rankings are according to this score.
\section{Results}
\begin{table*}[h]
 \centering
 \begin{tabular}{|c|c|c|c|}
 \hline
 Team Name & A+B+C & A\textsubscript{GT}+B+C & A\textsubscript{GT}+B\textsubscript{GT}+C\\
 \hline
 BioNLP@UIUC & 0.3828 & \textbf{0.7612} & \textbf{0.8594}\\
 ecnuica & 0.3335 & 0.7113 & 0.8145\\
 ITNLP & \textbf{0.4703} & 0.6863 & 0.7931\\
\textbf{KnowGraph@IITK} & \textbf{0.3783} & \textbf{0.6318} & \textbf{0.7600}\\
 INNOVATORS & 0.3205 & 0.5252 & 0.5971\\
 DuluthGrad & 0.2838 & 0.4921 & 0.7579\\
 YNU-HPCC & - & 0.4562 & 0.6541\\
 DFKI-SLT & 0.2651 & - & 0.7137\\
 NLP\_IITGN & - & 0.3522 & -\\
\hline
 \end{tabular}
 \caption{Average F1 score on test set of participating teams in end-to-end pipeline testing(A+B+C), phrase extraction testing(A\textsubscript{GT}+B+C) and triplet extraction testing(A\textsubscript{GT}+B\textsubscript{GT}+C).}
\label{table:result_1}
\end{table*}
Table \ref{table:result_1} shows all the participating teams average F1 score. In end-to-end pipeline testing, the input is the documents. In phrase extraction testing (A\textsubscript{GT}+B+C), the input is the document and ground-truth for sub-task A. Here, A\textsubscript{GT} represents the true label of sub-task A, and the F1 score for sub-task A is 1.00.  In the triplet extraction phase (A\textsubscript{GT}+B\textsubscript{GT}+C), the input is the document and ground truth labels (F1 score = 1.00) for both sub-tasks A and B to the system. Our team ``KnowGraph@IITK" achieved an F1 score of 0.3783 and ranked third in end-to-end pipeline testing. In phrase extraction and triplet extraction testing phases, our system ranked fourth with an F1 score of 0.6318 and 0.76, respectively. Our heuristic-based triplet extraction for Code and Research problem information unit achieved excellent performance with an F1 score of 1.00 and 0.9756, respectively, on the test set. 
\subsection{Ablation and Error Analysis}
\begin {table}[h]
\centering
\begin{tabular}{cc}
\hline
Methods & Dev F1 score\\
\hline
\multicolumn{2}{c}{Sub-task A}\\
\hline
SciBERT + CNN & 0.440\\
SciBERT + BiLSTM & \textbf{0.451} \\
\hline
\multicolumn{2}{c}{Sub-Task B}\\
\hline
BiLSTM + CRF & 0.361 \\
SciBERT + CRF & 0.444 \\
SciBERT + BiLSTM + CRF & \textbf{0.480}\\
\hline
\end{tabular}
\caption{F1 score of several methods of sub-task A and sub-task B on development set.}
\label{table:ablation_1}
\end {table}
We built several models using SciBERT, LSTM and CNN for each sub-task to understand each method's significance (Table \ref{table:ablation_1}). We found that language models are knowledge-rich and boost the existing models. On top of language models, BiLSTM based model performs better than the CNN-based model due to long semantic dependency in sequential models. In sub-task B, BiLSTM+CRF based model performed inferior to the same model built on top of SciBERT. In triplet formation in sub-task C, our rule-based approach of Research problem and Code information unit yield excellent results (highest on the leaderboard). A significant improvement in the F1 score for Baseline and Ablation Analysis IU suggests that the rule-based approach can boost neural models since specific patterns are present for these information units' triplets. In sub-task A, the dataset is highly skewed between minority and majority classes (1:10), making the training of a neural model difficult. On visualization of sub-task B outputs, we found some ambiguous phrases that our model fails to predict correctly. Extracting both scientific and relation cue phrases with high precision and recall in a single model is difficult. Sometimes, our model predicts scientific phrases correctly but fails to predict relation cue phrase in the same sentence.

In sub-task C, some IU triplet's such as Model, Hyperparameters, Results were present in large number comparatively and while Tasks and Dataset IU triplets are scarce in number. This skewness resulted in biasing of our multi-label classification model for IU prediction. In triplet formation, our predicate approach fails when the predicate is not present in the sentence and selected from the organizer's closed set of predicates.
 The proposed system also fails in the case of branching between the triplets, i.e. multiple triplets share the same subject phrase. Our triplet formation is unable to predict Approach, Dataset and Tasks triple. On visualizing, we found that Model triplets are predicted instead of Approach triplets. Further, the triplets lack enough features to be classified into an information unit using a neural-based method. Our system performance on sub-task C is poor in the end-to-end pipeline and phrase extraction testing because we submitted our initial experimentation model.
\section{Conclusion}
In this paper, we have presented our system for NLPContributionGraph task of SemEval 2021. We found that neural models combined with heuristics can build a knowledge graph by dividing it into small tasks. The heuristic-based model can outperform the neural approach in the triplet formation of some Information Units. In future work, neural models can use sparse transformers to encode long documents without increasing much memory. The pre-defined predicate incorporation could also be a future direction of work for our system.

\section*{Acknowledgments} 
We thank Shubham Kumar Nigam for his guidance and SemEval-2021 organizers for running the competition. We are also very grateful to the Pytorch developers. 
\bibliographystyle{acl_natbib}
\bibliography{acl2021}
\end{document}